\documentclass[10pt,twocolumn,letterpaper]{article}

\usepackage{cvpr}
\usepackage{times}
\usepackage{epsfig}
\usepackage{graphicx}
\usepackage{amsmath}
\usepackage{amssymb}

\usepackage[ruled,vlined]{algorithm2e}
\usepackage{enumerate}
\usepackage{subfigure}
\usepackage{graphicx}
\usepackage{array}
\usepackage{multirow}
\usepackage{comment}

\SetKwInput{kwInput}{Input}
\SetKwInput{kwOutput}{Output}
\SetKwInput{kwParameter}{Parameter}
\SetKwInput{kwIntermediate}{Intermediate Data}

\usepackage[pagebackref=true,breaklinks=true,letterpaper=true,colorlinks,bookmarks=false]{hyperref}

\cvprfinalcopy 

\ifcvprfinal\fi
\begin{document}

\title{Design of Efficient Convolutional Layers using Single Intra-channel Convolution, Topological Subdivisioning and Spatial ``Bottleneck'' Structure}


\author{
  Min Wang \\
  Department of EECS\\
  University of Central Florida\\
  Orlando, FL 32816\\
  \texttt{mwang@eecs.ucf.edu} \\
  \and
  Baoyuan Liu \\
  Department of EECS\\
  University of Central Florida\\
  Orlando, FL 32816\\
  \texttt{bliu@eecs.ucf.edu} \\
  \and
  Hassan Foroosh \\
  Department of EECS\\
  University of Central Florida\\
  Orlando, FL 32816\\
  \texttt{foroosh@eecs.ucf.edu} \\
}

\maketitle

\addtolength{\textfloatsep}{-1em}
\renewcommand{\baselinestretch}{0.97}
\selectfont

\begin{abstract}
   Deep convolutional neural networks achieve remarkable visual recognition performance, at the cost of high computational complexity. In this paper, we have a new design of efficient convolutional layers based on three schemes. 
The 3D convolution operation in a convolutional layer can be considered as performing spatial convolution in each channel and linear projection across channels simultaneously. By unravelling them and arranging the spatial convolution sequentially, the proposed layer is composed of a single intra-channel convolution, of which the computation is negligible, and a linear channel projection. A topological subdivisioning is adopted to reduce the connection between the input channels and output channels. Additionally, we also introduce a spatial ``bottleneck'' structure that utilizes a convolution-projection-deconvolution pipeline to take advantage of the correlation between adjacent pixels in the input. Our experiments demonstrate that the proposed  layers remarkably outperform the standard convolutional layers with regard to accuracy/complexity ratio. Our models achieve  similar accuracy to VGG, ResNet-50, ResNet-101 while requiring \textbf{42}, \textbf{4.5}, \textbf{6.5} times less computation respectively. 
 \end{abstract}

\section{Introduction}
Deep convolutional neural networks (CNN) have made significant improvement on solving visual recognition problems since the famous work by Krizhevsky \emph{et al.} in 2012 \cite{alexnet}\cite{googlenet}\cite{batchnorm}\cite{inception-v2}\cite{residual}. Thanks to their deep structure, vision oriented layer designs, and efficient training schemes, recent CNN models from Google \cite{inception-v2} and MSRA \cite{residual} obtain better than human level accuracy on ImageNet ILSVRC dataset \cite{imagenet}. 

The computational complexity for the state-of-the-art models for both training and inference are extremely high, requiring several GPUs or cluster of CPUs. The most time-consuming building block of the CNN, the convolutional layer, is performed by convolving the 3D input data with a series of 3D kernels. The computational complexity is quadratic in both the kernel size and the number of channels. To achieve state-of-the-art performance, the number of channels needs to be a few hundred, especially for the layers with smaller spatial input dimension, and the kernel size is generally no less than $3$.

Several attempts have been made to reduce the amount of computation and parameters in both convolutional layers and fully connected layers. Low rank decomposition has been extensively explored in various fashions \cite{low-rank-fergus}\cite{low-rank-vedaldi}\cite{low-rank1}\cite{low-rank2}\cite{low-rank3} to obtain moderate efficiency improvement. Sparse decomposition based methods \cite{scnn}\cite{deep-compress} achieve higher theoretical reduction of complexity, while the actual speedup is bounded by the efficiency of sparse multiplication implementations. Most of these decomposition-based methods start from a pre-trained model, and perform decomposition and fine-tuning based on it, while trying to maintain similar accuracy. This essentially precludes the option of improving efficiency by designing and training new CNN models from scratch.

On the other hand, in recent state-of-the-art deep CNN models, several heuristics are adopted to alleviate the burden of heavy computation. In \cite{googlenet}, the number of channels are reduced by a linear projection before the actual convolutional layer; In \cite{residual}, the authors utilize a bottleneck structure, in which both the input and the output channels are reduced by linear projection; In \cite{inception-v2}, $1 \times n$ and $n \times 1$ asymmetric convolutions are adopted to achieve larger kernel sizes. While these strategies to some extent help to design moderately efficient and deep models in practice, they are not able to provide a comprehensive analysis of optimizing the efficiency of the convolutional layer.

In this work, we propose several schemes to improve the efficiency of convolutional layers. In standard convolutional layers, the 3D convolution can be considered as performing intra-channel spatial convolution and linear channel projection simultaneously, leading to highly redundant computation. These two operations are first unraveled to a set of 2D convolutions in each channel and subsequent linear channel projection. Then, we make the further modification of performing the 2D convolutions sequentially rather than in parallel. In this way, we obtain a single intra-channel convolutional (SIC) layer that involves only one filter for each input channel and linear channel projection, thus achieving significantly reduced complexity. By stacking multiple  SIC layers, we can train models that are several times more efficient with similar or higher accuracy than models based on standard convolutional layer.

In a SIC layer, linear channel projection consumes the majority of the computation. To reduce its complexity, we propose a topological subdivisioning framework between the input channels and output channels as follows: The input channels and the output channels are first rearranged into a $s$-dimensional tensor, then each output channel is only connected to the input channels that are within its local neighborhood.  Such a framework leads to a regular sparsity pattern of the convolutional kernels, which is shown to possess a better performance/cost ratio than standard convolutional layer in our experiments. 

Furthermore, we design a spatial ``bottleneck'' structure to take advantage of the local correlation of adjacent pixels in the input. The spatial dimensions are first reduced by intra-channel convolution with stride, then recovered by deconvolution with the same stride after linear channel projection.  Such a design reduces the complexity of linear channel projection without sacrificing the spatial resolution. 

The above three schemes (SIC layer, topological subdivisioning and spatial ``bottleneck'' structure) attempt to improve the efficiency of traditional CNN models from different perspectives, and can be easily combined together to achieve lower complexity as demonstrated thoroughly in the remainder of this paper. Each of these schemes will be explained in detail in Section \ref{sec:method}, evaluated against traditional CNN models, and analyzed in Section \ref{sec:exp}.
 
\section{Method}
\label{sec:method}

\begin{figure}[t]
\centering
\subfigure[Standard Convolutional Layer] {
  \includegraphics[width=0.95\linewidth ]{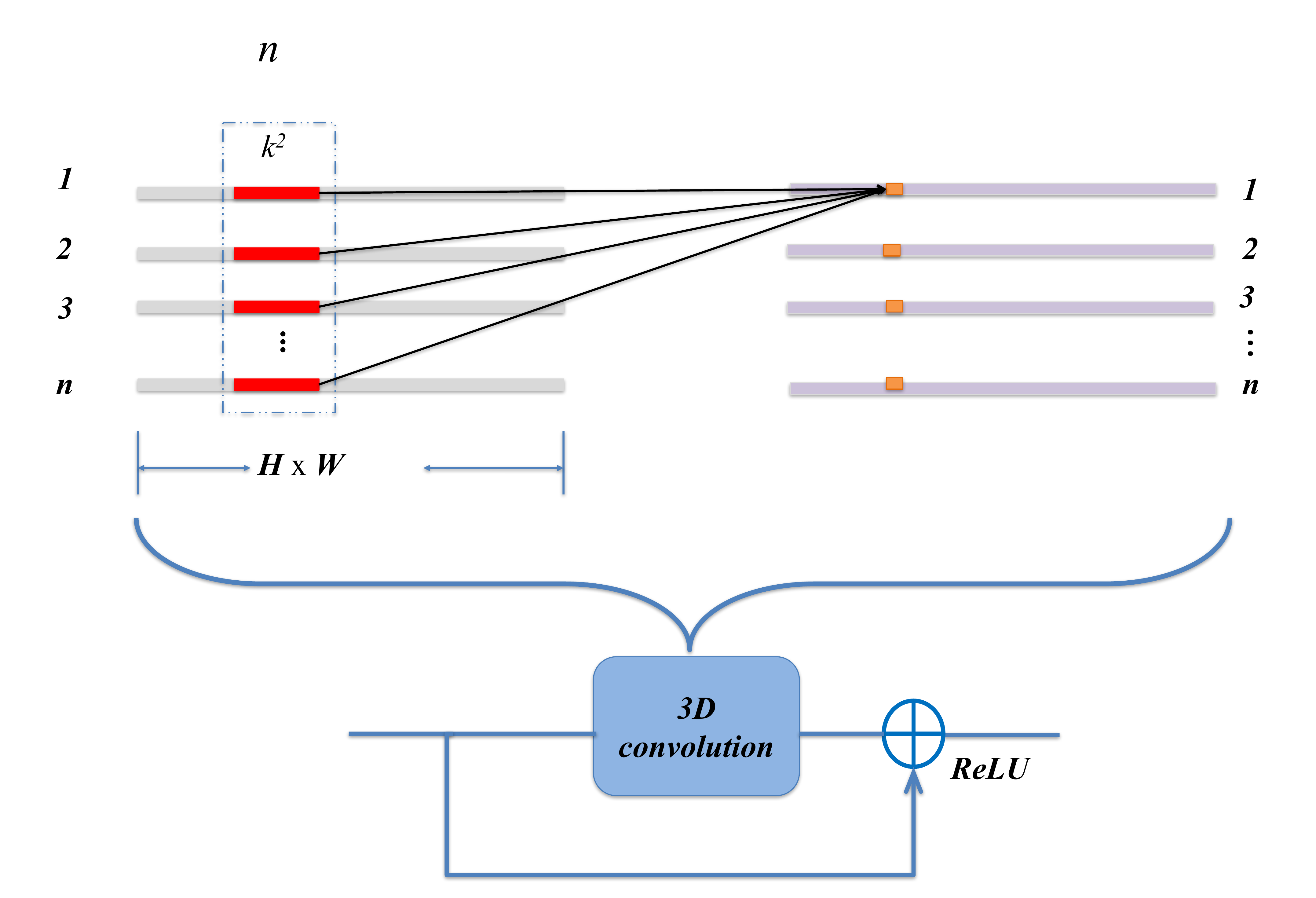}
  \label{fig: conv-standard}
}
\subfigure[Single Intra-Channel Convolutional Layer] {
  \includegraphics[width=0.95\linewidth]{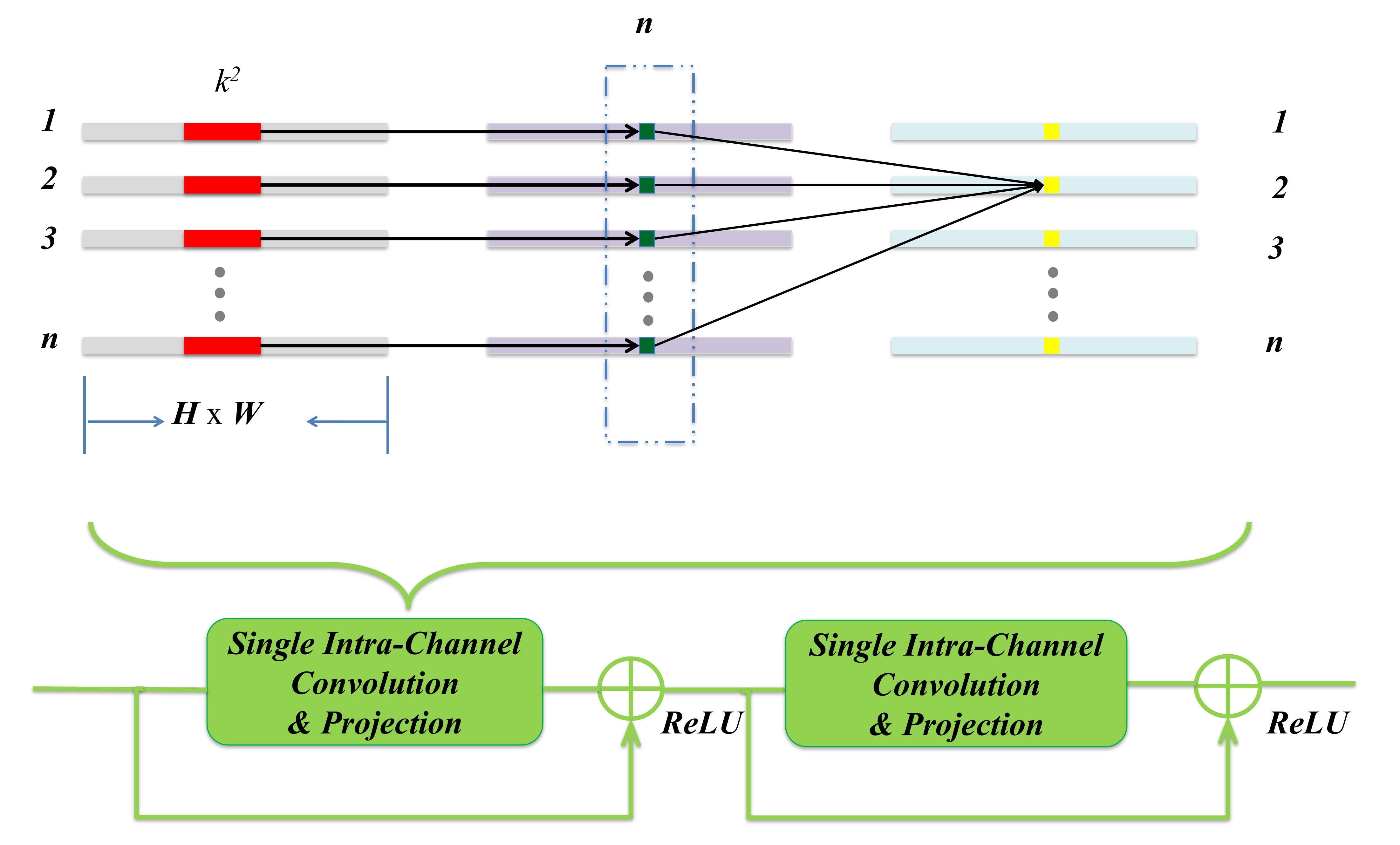}
  \label{fig: conf-single}
}
\caption{Illustration of the convolution pipeline of standard convolutional layer and Single Intra-channel Convolutional Layer. In SIC layer, only one 2D filter is convolved with each input channel.}
\label{fig: visualization}
\end{figure}

In this section, we first review the standard convolutional layer, then introduce the proposed schemes. For the purpose of easy understanding, the first two schemes are explained with mathematical equations and pseudo-code, as well as illustrated with graphical visualization in Figure \ref{fig: visualization}. 

We make the assumption that the number of output channels is equal to the number of input channels, and the input is padded so that the spatial dimensions of output is the same as input. We also assume that the residual learning technique is applied to each convolutional layer, namely the input is directly added to the output since they have the same dimension.

\subsection{Standard Convolutional Layer}
Consider the input data $\mathbf{I}$ in $\mathbb{R}^{h \times w \times n}$, where $h$, $w$ and $n$ are the height, width and the number of channels of the input feature maps, and the convolutional kernel $\mathbf{K}$ in $\mathbb{R}^{k \times k \times n \times n}$, where $k$ is size of the convolutional kernel and $n$ is the number of output channels. The operation of a standard convolutional layer $\mathbf{O} \in \mathbb{R}^{h \times w \times n} =  \mathbf{K}*\mathbf{I}$ is given by Algorithm \ref{alg: standard-conv}. The complexity of a convolutional layer measured by the number of multiplications is

\begin{equation}
n^2k^2hw
\end{equation}

Since the complexity is quadratic with the kernel size, in most recent CNN models, the kernel size is limited to $3 \times 3$ to control the overall running time.  

\begin{algorithm}[h]
\label{alg: standard-conv}
\SetAlgoLined
\DontPrintSemicolon
\caption{Standard Convolutional Layer}
\kwInput{$\mathbf{I} \in \mathbb{R}^{h \times w \times n}$}
\kwParameter{$\mathbf{K} \in \mathbb{R}^{k \times k \times n \times n}$}
\kwIntermediate{$\mathbf{\tilde{I}} \in \mathbb{R}^{(h + k - 1) \times (w + k - 1) \times n}$}
\kwOutput{$\mathbf{O} \in \mathbb{R}^{h \times w \times n}$}
$\mathbf{\tilde{I}} = \text{zero-padding}(\mathbf{I}, \frac{k - 1}{2})$\;
\For{$y = 1 ~ \KwTo ~ h, ~ x = 1 ~ \KwTo ~ w, ~ j = 1 ~ \KwTo ~ n$} {
  $\mathbf{O}(y,x,j)=\displaystyle\sum_{i=1}^{n}\sum_{u=1}^{k}\sum_{v=1}^{k}\mathbf{K}(u,v,i,j)\mathbf{\tilde{I}}(y+u-1,x+v-1,i)$
}
\end{algorithm}

\subsection{Single Intra-Channel Convolutional Layer}
\label{sec: sic}

In standard convolutional layers, the output features are produced by convolving  a group of 3D kernels with the input features along the spatial dimensions. Such a 3D convolution operation can be considered as a combination of 2D spatial convolution inside each channel and linear projection across channels. For each output channel, a spatial convolution is performed on each input channel. The spatial convolution is able to capture local structural information, while the linear projection transforms the feature space for learning the necessary non-linearity in the neuron layers. When the number of input and output channels is large, typically hundreds, such a 3D convolutional layer requires an exorbitant amount of computation.

A natural idea is, the 2D spatial convolution and linear channel projection can be unraveled and performed separately. Each input channel is first convolved with $b$ 2D filters, generating intermediate features that have $b$ times channels of the input. Then the output is generated by linear channel projection. Unravelling these two operations provides us more freedom of model design by tuning both $b$ and $k$.  
The complexity of such a layer is 
\begin{equation}
b(nk^2 + n^2)hw
\end{equation}

Typically, $k$ is much smaller than $n$. The complexity is approximately linear with $b$. 
When $b = k^2$, this is equivalent to a linear decomposition of the standard convolutional layers \cite{scnn}. When $b < k^2$, the complexity is lower than the standard convolutional layer in a low-rank fashion.




Our key observation is that instead of convolving $b$ 2D filters with each input channel simultaneously, we can perform the convolutions sequentially. The above convolutional layer with $b$ filters can be transformed to a framework that has $b$ layers. In each layer, each input channel is first convolved with \textbf{single} 2D filter, then a linear projection is applied to all the input channels to generate the output channels. In this way, the number of channels are maintained the same throughout all $b$ layers. Algorithm. \ref{alg: sic} formally describes this framework. 

When we consider each of the $b$ layers, only one $k \times k$ kernel is convolved with each input channel. This seems to be a risky choice. Convolving with only one filter will not be able to preserve all the information from the input data, and there is very little freedom to learn all the useful local structures. Actually, this will probably lead to a low pass filter, which is somewhat equivalent to the first principal component of the image. However, the existence of residual learning module helps to overcome this disadvantage. With residual learning, the input is added to the output. The subsequent layers thus receive information from both the initial input and the output of preceding layers. Figure. \ref{fig: visualization} presents a visual comparison between the proposed method and standard convolutional layer. 

\begin{algorithm}[h]
\label{alg: sic}
\SetAlgoLined
\DontPrintSemicolon
\caption{Single Intra-Channel Convolutional Layer}
\kwInput{$\mathbf{I} \in \mathbb{R}^{h \times w \times n}$}
\kwParameter{$\mathbf{K} \in \mathbb{R}^{k \times k  \times n}, \mathbf{P} \in \mathbb{R}^{ n \times n}$}
\kwIntermediate{$\mathbf{\tilde{I}} \in \mathbb{R}^{(h + k - 1) \times (w + k - 1) \times n}$, $\mathbf{G} \in \mathbb{R}^{h \times w\times n}$}
\kwOutput{$\mathbf{O} \in \mathbb{R}^{h \times w \times n}$}
$\mathbf{O} = \mathbf{I}$ \tcp*[l]{Initialize output as input}
$\mathbf{\tilde{I}} = \text{zero-padding}(\mathbf{I}, \frac{k - 1}{2})$\;
\For(\tcp*[h]{Repeat this layer $b$ times}){$i = 1 ~ \KwTo ~ b$} {
\For{$y = 1 ~ \KwTo ~ h, ~ x = 1 ~ \KwTo ~ w, ~ j = 1 ~ \KwTo ~ n$} {
  $\mathbf{G}(y,x,j)=\displaystyle\sum_{u=1}^{k}\sum_{v=1}^{k}\mathbf{K}(u,v,j)\mathbf{\tilde{I}}(y+u-1,x+v-1,j)$
}
\For{$y = 1 ~ \KwTo ~ h, ~ x = 1 ~ \KwTo ~ w, ~ l = 1 ~ \KwTo ~ n$} {
  $\mathbf{O}(y,x,l) = \mathbf{O}(y,x,l) + \displaystyle\sum_{j=1}^{n}\mathbf{G}(y,x,j)\mathbf{P}(j, l)$
}
$\mathbf{O} = \max(\mathbf{O}, 0)$ \tcp*[l]{ReLU}
$\mathbf{\tilde{I}} = \text{zero-padding}(\mathbf{O}, \frac{k - 1}{2})$\;
}
\end{algorithm}

\subsection{Topologica Subdivisioning}
Given that the standard convolutional layer boils down to single intra-channel convolution and linear projection in the SIC layer, we make further attempt to reduce the complexity of linear projection. In \cite{scnn}, the authors proved that extremely high sparsity could be accomplished without sacrificing accuracy. While the sparsity was obtained by fine-tuning and did not possess any structure, we study to build the sparsity with more regularity. Inspired by the topological ICA framework in \cite{tica}, we propose a $s$-dimensional topological subdivisioning between the input and output channels in the convolutional layers. Assuming the number of input channels and output channels are both $n$, we first arrange the input and output channels as an $s$-dimensional tensor $[d_1, d_2, ..., d_s]$.
\begin{equation}
\prod_{i=1}^{s}d_i  = n;
\end{equation}
Each output channel is only connected to its local neighbors in the tensor space rather than all input channels. The size of the local neighborhood is defined by another $s$-dimensional tensor, $[c_1, c_2, ..., c_s]$, and the total number of neighbors for each output channel is 
\begin{equation}
\prod_{i=1}^{s}c_i  = c;
\end{equation}

\begin{figure}[t]
\centering
\subfigure[2D Topology] {
  \includegraphics[width=0.75\linewidth ]{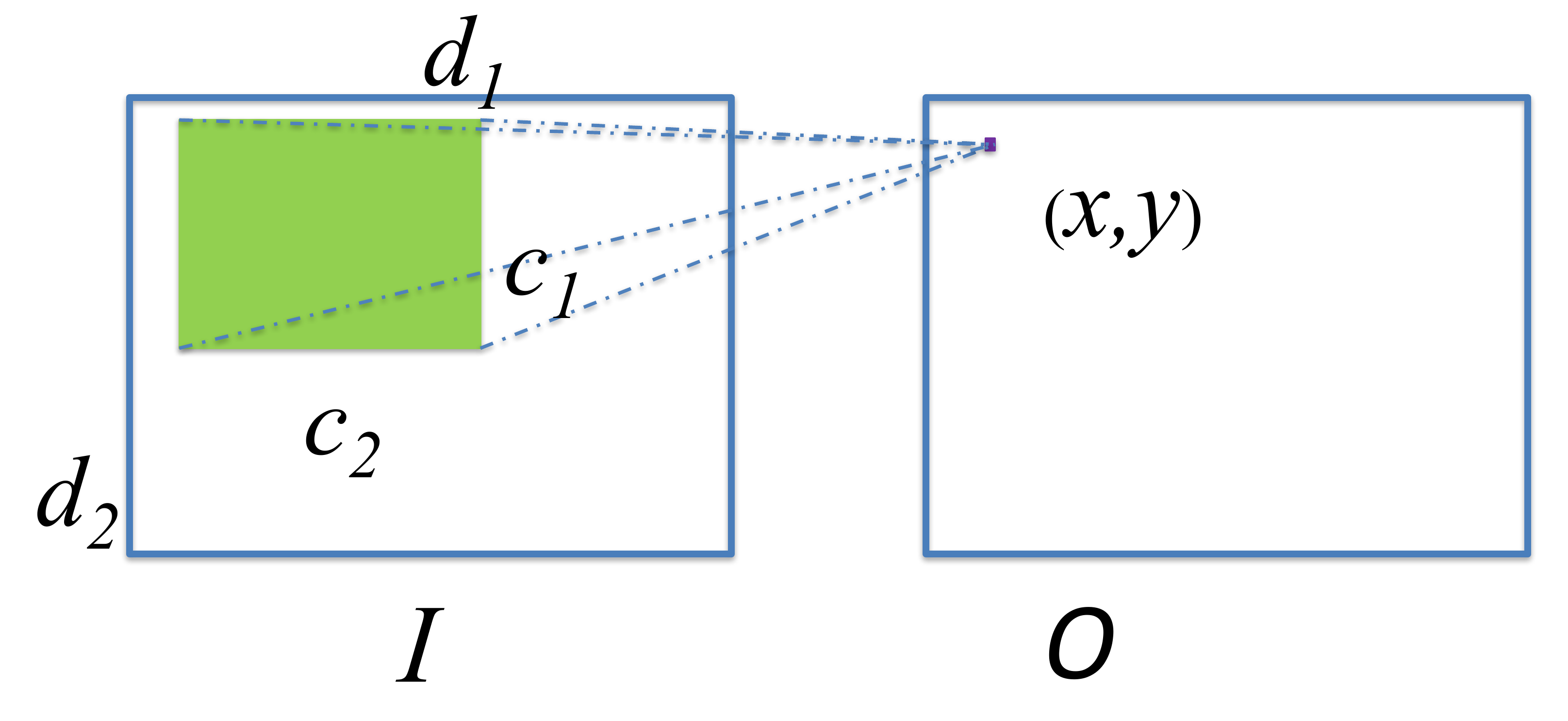}
  \label{fig: 2d-top}
}
\subfigure[3D Topology] {
  \includegraphics[width=0.75\linewidth]{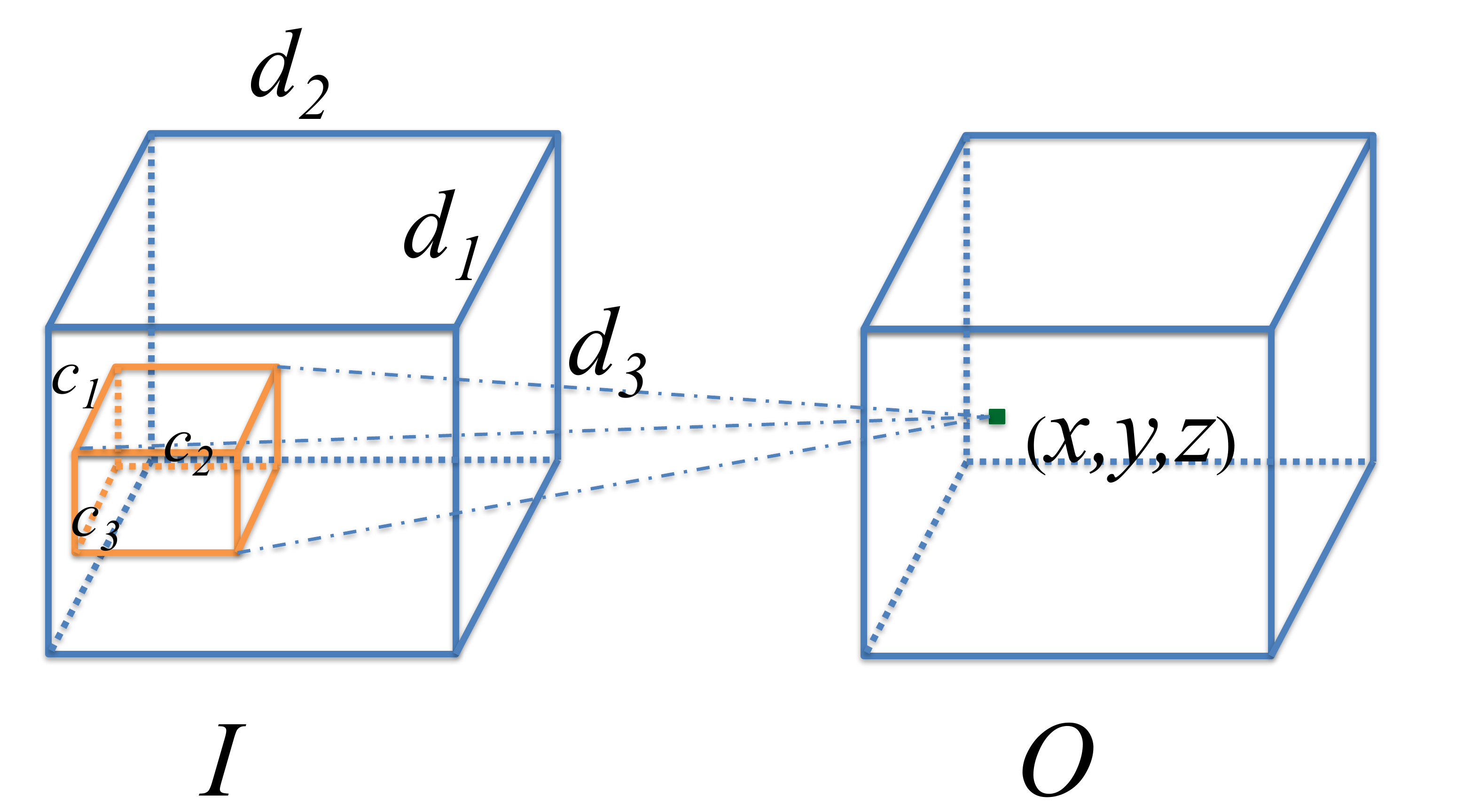}
  \label{fig: 3d-top}
}
 \caption{ 2D \&3D topology for input and output. }
 \label{fig: topology}
\end{figure}

\begin{algorithm}[b]
\label{alg: topology}
\SetAlgoLined
\DontPrintSemicolon
\caption{Convolutional Layer with Topological Subdivisioning}
\kwInput{$\mathbf{I} \in \mathbb{R}^{h \times w \times n}$}
\kwParameter{$\prod_{i=1}^{s}d_i = n; c_i \leq d_i, \forall i = 1 ... s; \mathbf{K} \in \mathbb{R}^{k \times k \times d_1 \times .. \times d_s \times c_1 \times ... \times c_s}$}
\kwIntermediate{$\mathbf{\tilde{I}} \in \mathbb{R}^{(h + k - 1) \times (w + k - 1) \times n}, \mathbf{\hat{I}} \in \mathbb{R}^{(h + k - 1) \times (w + k - 1) \times d_1 \times ... \times d_s}$}
\kwOutput{$\mathbf{O} \in \mathbb{R}^{h \times w \times d_1 \times ... \times d_s}$}
$\mathbf{\tilde{I}} = \text{zero-padding}(\mathbf{I}, \frac{k - 1}{2})$\;
Rearrange $\mathbf{\tilde{I}}$ to $\mathbf{\hat{I}}$\;
\For{$y = 1 ~ \KwTo ~ h, ~ x = 1 ~ \KwTo ~ w, ~ j_1 = 1 \KwTo ~ d_1, ... ~ j_s = 1 \KwTo ~ d_s$} {
  \tcp*[l]{Topological Subdivisioning}
  \begin{align*}
  \mathbf{O}(y,x,j_1, ..., j_s) ~~~~& \\
   = \displaystyle\sum_{i_1=1}^{c_1}...\sum_{i_s=1}^{c_s}\sum_{u,v=1}^{k} & \mathbf{K}(u,v,j_1, ..., j_s, i_1, ..., i_s) \cdot \\ 
   &\mathbf{\tilde{I}}(y+u-1,x+v-1, \\ 
   &(j_1 + i_1 - 2) \% d_1 + 1, \\
   &..., \\
   &(j_s + i_s - 2) \% d_s + 1)
  \end{align*}
}
\end{algorithm}

The complexity of the proposed topologically subdivisioned convolutional layers compared to the standard convolutional layers can be simply measured by $\frac{c}{n}$. Figure. \ref{fig: topology} illustrate the 2D and 3D  topological subdivisioning between the input channels and the output channels.  A formal description of this layer is presented in Algorithm \ref{alg: topology}.

When $k = 1$, the algorithm is suitable for the linear projection layer, and can be directly embedded into Algorithm \ref{alg: sic} to further reduce the complexity of the SIC layer. 

\subsection{Spatial ``Bottleneck'' Structure}

In the design of traditional CNN models, there has always been a trade-off between the spatial dimensions and the number of channels. While high spatial resolution is necessary to preserve detailed local information, large number of channels produce high dimensional feature spaces and learn more complex representations.The complexity of one convolutional layer is determined by the product of these two factors. To maintain an acceptable complexity, the spatial dimensions are reduced by max pooling or stride convolution while the number of channels are increased. 

\begin{figure}[t]
\centering
\includegraphics[width=0.8\linewidth]{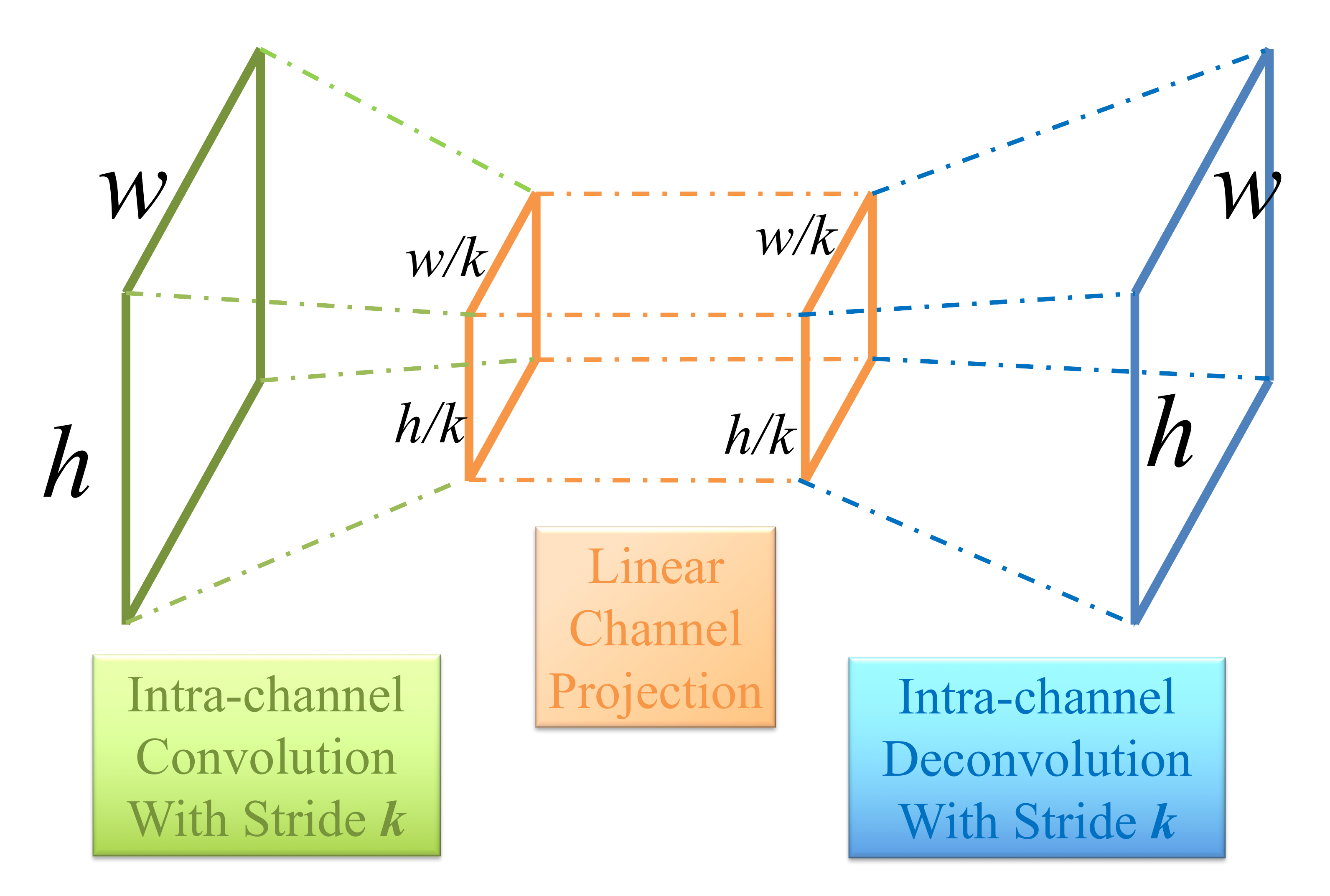}
\caption{Illustration of Spatial ``Bottleneck'' Framework}
\label{fig: spatial-bottleneck}
\end{figure} 

On the other hand, the adjacent pixels in the input of each convolutional layers are correlated, in a similar fashion to the image domain, especially when the spatial resolution is high. While reducing the resolution by simple sub-sampling will obviously lead to a loss of information, such correlation presents considerable redundancy that can be taken advantage of. 

In this section, we introduce a spatial ``bottleneck'' structure that reduces the amount of computation without decreasing either the spatial resolution or the number of channels by exploiting the spatial redundancy of the input.

Consider the 3D input data $\mathbf{I}$ in $\mathbb{R}^{h \times w \times n}$, we first apply a single intra-channel convolution to each input channel as was introduced in Section \ref{sec: sic}. A $k \times k$ kernel is convolved with each input channel with stride $k$, so that the output dimension is reduced to $\mathbb{R}^{\frac{h}{k} \times \frac{w}{k} \times n}$. Then a linear projection layer is applied. Finally, We perform a $k \times k$ intra-channel deconvolution with stride $k$ to recover the spatial resolution. 

\begin{table*}[t]
\begin{center}
\begin{tabular}[t]{c c c c c c c} \hline
Stage & Output &   \textbf{A} &  \textbf{B} &  \textbf{C}  & \textbf{D} & \textbf{E}\\ \hline
1& $108^2$ &  \multicolumn{5}{c}{$(7, 64)_2$} \\ \hline
\multirow{3}{*}{2 } & \multirow{3}{*}{$36^2 $} & \multicolumn{5}{c}{$3\times3$ max pooling , stride 3} \\ \cline{3-7}
& & \multicolumn{5}{c}{($1$, 128)} \\ \cline{3-7}
& & $(3, 128)\times 2$
& $[3, 4, 128]\times 2$  
& $<3, 128> \times 4$
& $<5, 128> \times 4$
& $<3, 128> \times 6$ \\	\hline				

\multirow{3}{*}{3 } & \multirow{3}{*}{$18 ^2$} &\multicolumn{5}{c}{$2\times2$ max pooling , stride 2} \\ \cline{3-7}
& & \multicolumn{5}{c}{($1$, 256)} \\ \cline{3-7}
& & $(3, 256)\times 2$
& $[3, 4, 256]\times 2$  
& $<3, 256> \times 4$
& $<5, 256> \times 4$
& $<3, 256> \times 6$ \\	\hline					
					
\multirow{3}{*}{4} & \multirow{3}{*}{$6^2$} & \multicolumn{5}{c}{$3\times3$ max pooling , stride 3} \\ \cline{3-7}
& & \multicolumn{5}{c}{($1$, 512)} \\ \cline{3-7}
& & $(3, 512)\times 2$
& $[3, 4, 512]\times 2$  
& $<3, 512> \times 4$
& $<5, 512> \times 4$
& $<3, 512> \times 6$ \\  \cline{3-7}				
& & \multicolumn{5}{c}{($1$, 1024)} \\ \hline
& \multirow{4}{*}{$1^2$} & \multicolumn{5}{c}{$6 \times 6$ average pooling, stride 6} \\
&  & \multicolumn{5}{c}{fully connected, 2048} \\ 
&  & \multicolumn{5}{c}{fully connected, 1000} \\ 
&  & \multicolumn{5}{c}{softmax} \\ \hline
\end{tabular}
\caption{Configurations of baseline models and models with proposed SIC layers . For each convolutional layer, we use numbers in brackets to represent its configuration. $k$ denotes the kernel size.  $n$ is the number of output channels. 
Different types of bracket correspond to different convolutional layer. $( k , n )$ is a typical standard convolutional layer. $[ k, b , n  ]$ denotes an unraveled convolutional layer with $b$ filters for each input channel. $<k, n>$ represents our SIC layer. The number after the brackets indicates the times that the layer is repeated in each stage.}
\label{tab:model-config}
\end{center}
\end{table*}

Figure. \ref{fig: spatial-bottleneck} illustrates the proposed spatial ``bottleneck'' framework. The spatial resolution of the data is first reduced, then expanded, forming a bottleneck structure. In this 3-phase structure, the linear projection phase , which consumes most of the computation, is $k^2$ times more efficient than plain linear projection on the original input. The intra-channel convolution and deconvolution phases learn to capture the local correlation of adjacent pixels, while maintaining the spatial resolution of the output. 

\section{Experiments}
\label{sec:exp}
We evaluate the performance of our method on the ImageNet LSVRC 2012 dataset, which contains 1000 categories, with 1.2M training images, 50K validation images, and 100K test images. We use Torch to train the CNN models in our framework. Our method is implemented with CUDA and Lua based on the Torch platform. The images are first resized to $256\times 256$, then randomly cropped into $221 \times 221$ and flipped horizontally while training. Batch normalization \cite{batchnorm} is placed after each convolutional layer and before the ReLU layer.
We also adopt the dropout \cite{dropout} strategy with a ratio of 0.2 during training. Standard stochastic gradient descent with mini-batch containing 256 images is used to train the model. We start the learning rate from 0.1 and divide it by a factor of 10 every 30 epochs. Each model is trained for 100 epochs. For batch normalization, we use exponential moving average to calculate the batch statistics as is implemented in CuDNN \cite{cudnn}. The code is run on a server with 4 Pascal Titan X GPU. For all the models evaluated below, the top-1 and top-5 error of validation set with central cropping is reported. 

\begin{table}[t]
\begin{center}
\begin{tabular}{c c  c  c  c c } \hline
Stage & 2 & 3 & 4 \\ \hline
Intra-channel Convolution & 6.6\% & 3.4\% & 1.7\% \\
Linear Projection & 93.4\% & 96.6\% & 98.3\% \\ \hline
\end{tabular}
\end{center}
\caption{Distribution of the computation in the SIC layer of Model \textbf{C}. The intra-channel convolution generally consumes less than 10\% of total computation, and its proportion decreases when the number of channels increases.}
\label{tab:distribution}
\end{table}

We evaluate the performance and efficiency of a series of models designed using the proposed efficient convolutional layer. To make cross reference easier and help the readers keep track of all the models, each model is indexed with a capital letter. 

\begin{table*}[t]
\begin{center}
\begin{tabular}{c c  c  c  c c} \hline
Model &  kernel size & \# layers& Top-1 err. & Top-5 err. & Complexity  \\ \hline
\textbf{A}  & 3&2&  30.67\% &11.24\% & 1 \\
\textbf{B}  & 3&2 & 30.69\% &11.27\%& \textasciitilde 4/9 \\
\textbf{C}  & 3&4 & 29.78\%&10.78\% & \textasciitilde 2/9 \\
\textbf{D}  & 5&4& 29.23\% &10.48\%& \textasciitilde  2/9 \\ 
\textbf{E}  & 3&6& 28.83\% &9.88\%& \textasciitilde 1/3 \\ \hline
\end{tabular}
\end{center}
\caption{Top-1 \& Top-5 error and complexity per stage of model \textbf{A} to \textbf{E}. The models with proposed design (model \textbf{C}, \textbf{D}, \textbf{E})demonstrate significantly better accuracy / complexity ratio than the baseline model.}
\label{tab:acc-conv-bases}
\end{table*}

We compare our method with a baseline CNN model that is built from standard convolutional layers. The details of the baseline models are given in Table \ref{tab:model-config}. The convolutional layers are divided into stages according to their spatial dimensions. Inside each stage, the convolutional kernels are performed with paddings so that the output has the same spatial dimensions as the input. Across the stages, the spatial dimensions are reduced by max pooling and the number of channels are doubled by $1\times 1$ convolutional layer. One fully connected layer with dropout is added before the logistic regression layer for final classification. 
Residual learning is added after every convolutional layer with same number of input and output channels.

We evaluate the performance of our method by substituting the standard convolutional layers in the baseline models with the proposed Single Intra-Channel Convolutional (SIC) layers. We leave the $7\times 7$ convolutional layer in the first stage and the $1 \times 1$ convolutional layers across stages the same, and only substitute the $3 \times 3$ convolutional layers. In the following sections, the relative complexities are also measured with regards to these layers.

\subsection{Single Intra-Channel Convolutional Layer}

We first substitute the standard convolutional layer with the unraveled convolution configuration in model \textbf{B}. Each input channel is convolved with 4 filters, so that the complexity of \textbf{B} is approximately $\frac{4}{9}$ of the baseline model \textbf{A}. In model \textbf{C} , we use two SIC layers to replace one standard convolutional layer. Even though our model \textbf{C} has more layers than the baseline model \text{A}, its complexity is only $\frac{2}{9}$ of model \textbf{A}. In model \textbf{E}, we increase the number of SIC layers from 4 in model \textbf{C} to 6 in model \textbf{E}. The complexity of model \textbf{E} is only $\frac{1}{3}$ of the baseline. Due to the extremely low complexity of the SIC layer, we can easily increase the model depth without too much increase of the computation. 

Table. \ref{tab:distribution} lists the distribution of computation between the intra-channel convolution and linear channel projection of each SIC layer in model \textbf{C}. The intra-channel convolution generally consumes less than 10\% of the total layer computation. Thanks to this advantage, we can utilize a larger kernel size with only a small sacrifice of efficiency. Model \textbf{D} is obtained by setting the kernel size of model \text{C} to 5.

Table \ref{tab:acc-conv-bases} lists the top-1 and top-5 errors and the complexity of models from \textbf{A} to \textbf{E}. Comparing model \textbf{B} and \textbf{A}, with same number of layers, model \textbf{B} can match the accuracy of model \textbf{A} with less than half computation. When comparing the SIC based model \textbf{C} with model \textbf{B},  model \textbf{C} reduces the top-1 error by 1\% with half complexity. This verifies the superior efficiency of the proposed SIC layer. With $5 \times 5$ kernels, model \textbf{E} obtains 0.5\% accuracy gain with as low as 5\% increase of complexity on average. This demonstrates that increasing kernel size in SIC layer provides us another choice of improving the accuracy/complexity ratio. 

\subsection{Topological Subdivisioning}
We first compare the performance of two different topological configurations against the baseline model. Model \textbf{F} adopts 2D topology and $c_i = d_i/2$ for both dimensions, which leads to a reduction of complexity by a factor of 4. In Model \textbf{G}, we use 3D topology and set $c_i$ and $d_i$, so that the complexity is reduced by a factor of 4.27. The details of the network configuration are listed in Table \ref{tab: topology-config}. The number of topological layers is twice the number of standard convolutional layers in the baseline model, so the overall complexity per stage is reduced by a factor of 2.

\begin{table}[h]
\begin{center}
\begin{tabular}{c c c c} \hline
\multirow{3}{*}{Stage} & \multirow{3}{*}{\#Channels} & 2D topology & 3D topology \\ 
&   & $d_1 \times d_2$ & $d_1 \times d_2 \times d_3$ \\
& & $c_1 \times c_2$ & $c_1 \times c_2 \times c_3$ \\ \hline
 \multirow{2}{*}{2}& \multirow{2}{*}{128 }& $8 \times 16$ & $ 4\times8 \times 4$ \\ 
& & $4\times 8$ & $2\times 5 \times 3$ \\ \hline
\multirow{2}{*}{3}& \multirow{2}{*}{256 }& $16\times 16$ & $8 \times 8 \times 4$ \\
& & $8 \times 8$ & $4 \times 5 \times 3$ \\ \hline
\multirow{2}{*}{4}& \multirow{2}{*}{512 }& $16 \times 32$ & $8 \times 8 \times 8$ \\ 
& & $8 \times 16$ & $4 \times 5 \times 6$ \\ \hline
\end{tabular}
\end{center}
\caption{Configurations of model \textbf{F} and \textbf{G} that use 2D and 3D topological subdivisioning. $d_i$ and $c_i$ stand for the tensor and neighbor dimensions in Algorithm \ref{alg: topology}. They are designed so that the complexity is reduced by (approximately for 3D) a factor of 4. }
\label{tab: topology-config}
\end{table}

As a comparison, we also train a model \textbf{H} using the straightforward grouping strategy introduced in \cite{alexnet}. Both the input and output channels are divided into 4 groups. The output channels in each group are only dependent on the input channels in the corresponding group. The complexity is also reduced 4 times in this manner. Table \ref{tab: topology-acc} lists the top-1 \& top-5 error rate and complexities of model \textbf{F} to \textbf{H}. Both the 2D and the 3D topology models outperform the grouping method with lower error rate while maintaining the same complexity. When compared with the baseline model, both of the two topology models achieve similar top-1 and top-5 error rate with half the computation. 
\begin{table}[t]
\begin{center}
\begin{tabular}{c c c c c } \hline
Model & Methods & Top-1 & Top-5 & Complexity  \\ \hline
\textbf{A} & Baseline & 30.67\% &11.24\%& 1  \\
\textbf{H} & Grouping & 31.23\%&11.73\% & \textasciitilde 1/2 \\ 
\textbf{F} & 2D Top & 30.53\%&11.28\% & \textasciitilde 1/2  \\
\textbf{G} & 3D Top & 30.69\%&11.38\%  & \textasciitilde 15/32  \\
\textbf{I} & SIC+2D & 30.78\% &11.29\%& \textasciitilde 1/9 \\ \hline
\end{tabular}
\caption{Top-1\&Top-5 error rate and complexity of topology models and grouping model.}
\label{tab: topology-acc}
\end{center}
\end{table}

Finally, we apply the topological subdivisioning to the SIC layer in model \textbf{I}. We choose 2D topology based on the results in Table \ref{tab: topology-acc}. In model \textbf{I}, there are 8 convolutional layers for each stage, due to the layer doubling caused by both the SIC layer and the topological subdivisioning. The complexity of each layer is, however, approximately as low as $\frac{1}{36}$ of a standard $3\times 3$ convolutional layer. Compared to the baseline model, 2D topology together with SIC layer achieves similar error rate while being 9 times faster.

\subsection{Spatial ``Bottleneck'' Structure}
In our evaluation of layers with spatial ``bottleneck'' structure, both the kernel size and the stride of the in-channel convolution and deconvolution is set to 2. The complexity of such a configuration is a quarter of a SIC layer. Both model \textbf{J} and model \textbf{K} are modified from model \textbf{C} by replacing SIC layers with spatial ``bottleneck'' layers. One SIC layer is substituted with two Spatial ``Bottleneck'' layers, the first one with no padding and the second one with one pixel padding, leading to a 50\% complexity reduction. In model \textbf{J}, every other SIC layer is substituted; In model \textbf{K}, all SIC layers are substituted. Table \ref{tab: spatial-acc} compares their performance with the baseline model and SIC based model. Compared to the SIC model \textbf{C}, model \textbf{J} reduces the complexity by 25\% with no loss of accuracy; model \textbf{K} reduces the complexity by 50\% with a slight drop of accuracy. Compared to the baseline model \textbf{A}, model \textbf{K} achieves 9 times speedup with similar accuracy.

\begin{table}[h]
\begin{center}
\begin{tabular}{c c c c c } \hline
Model & \#layers & Top-1 err. & Top-5 err.& Complexity  \\ \hline
\textbf{A} & 2 & 30.67\% &11.24\%& 1  \\
\textbf{C}& 4 & 29.78\%& 10.78\%&  \textasciitilde 2/9 \\
\textbf{J} & 6 & 29.72\% & 10.66\%& \textasciitilde 1/6 \\
\textbf{K} & 8& 30.78\% & 11. 34\%& \textasciitilde 1/9 \\ \hline
\end{tabular}
\caption{Top-1\&Top-5 error rate  and complexity of SIC layer with spatial ``bottleneck'' structure.}
\label{tab: spatial-acc}
\end{center}
\end{table}

\subsection{Comparison with standard CNN models}

In this section, we increase the depth of our models to compare with recent state-of-the-art CNN models. 
To go deeper but without increasing too much complexity, we adopt the channel-wise bottleneck structure similar to the one introduced in \cite{residual}. In each channel-wise bottleneck structure, the number of channels are first reduced by half by the first layer, then recovered by the second layer. Such a two-layer bottleneck structure has almost the same complexity to single layer with the same input and output channels, thus increase the overall depth of the network. 

\begin{figure}[t]
\centering
\includegraphics[width=0.9\linewidth]{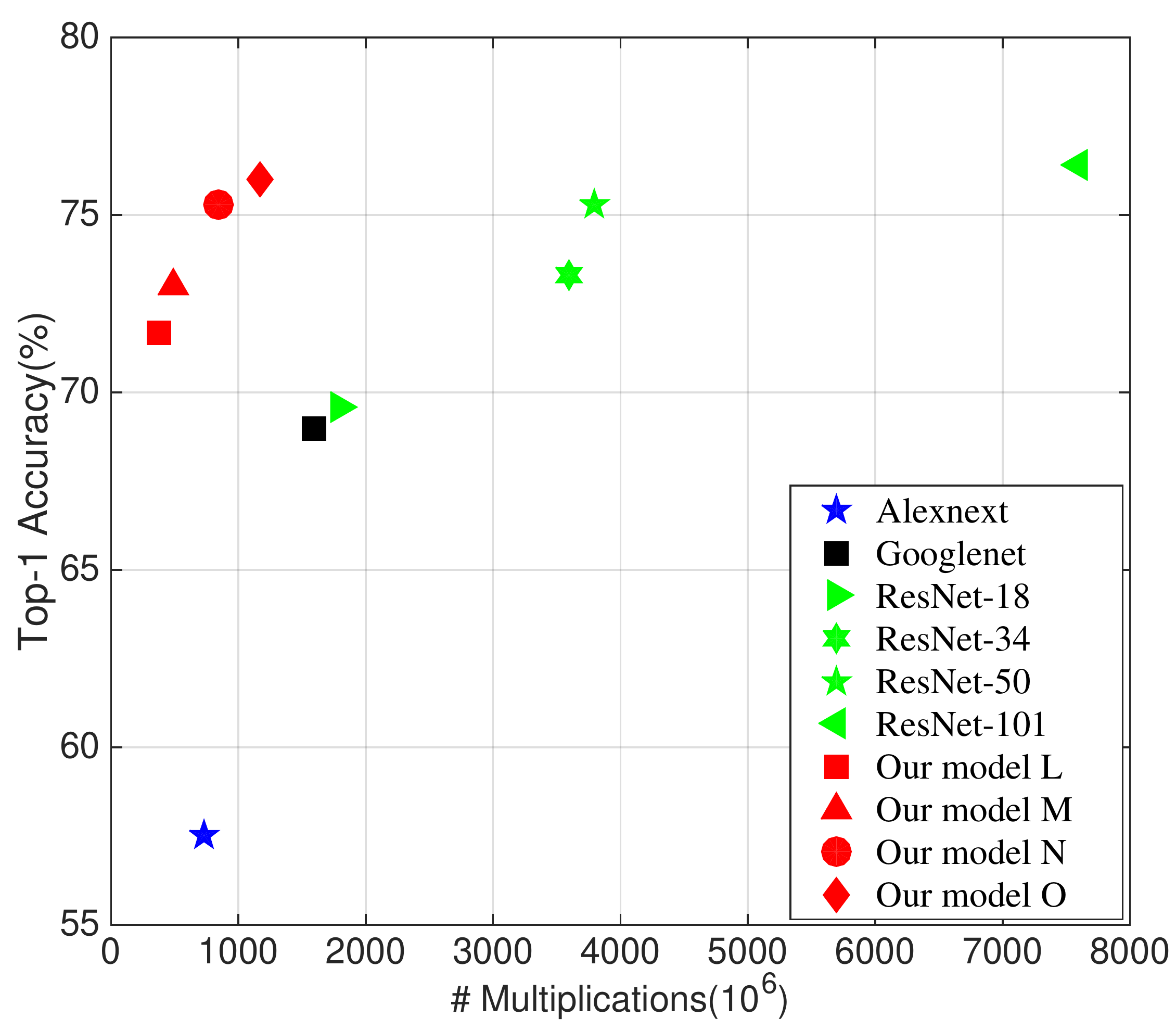}
\caption{Comparing top-1 accuracy and complexity between our model and several previous work}
\label{fig: scatter}
\end{figure}

We gradually increase the number of SIC layers with channel-wise bottleneck structure in each stage from 8 to 40, and compare their complexity to recent CNN models with similar accuracies. Model $\bf{L}$ , $\bf{M}$, $\bf{N}$ and $\bf{O}$ correspond to the number of layers of 8, 12, 24, and 40, respectively. Due to training memory limitation, only the SIC layer is used in models in this section. While model $\bf{L}$ and $\bf{M}$ have the same spatial dimensions and stage structures as in Table \ref{tab:model-config}, model $\bf{N}$ and $\bf{O}$ adopt the same structure as in \cite{residual}. They have different pooling strides and one more stages right after the first $7\times 7$ convolutional layer. The detailed model configurations are put in the supplemental materials. 

\begin{table*}[t]
\begin{center}
\begin{tabular}[t]{c c  c  c  } \hline
Model &   Top-1 err. & Top-5 err. & \# of Multiplications\\ \hline
AlexNet & 42.5\%& 18.2\% & 725M \\
GoogleNet & 31.5\% &10.07\%& 1600M \\
ResNet\_18 & 30.43\% &10.76\%& 1800M \\
VGG &  28.5\%& 9.9\% &16000M \\
\bf{Our Model \textbf{L}} &\textbf{ 28.29\%} & \textbf{9.9\%}& \textbf{381M} \\ \hline
ResNet\_34 & 26.73\% & 8.74\% & 3600M \\
\bf{Our Model \textbf{M}} &\textbf{ 27.07\%}&\textbf{8.93\%}&\textbf{ 490M }\\ \hline
ResNet\_50 & 24.7\% & 7.8\% & 3800M \\
\bf{Our Model \textbf{N}} & \textbf{24.76\%}&\textbf{ 7.58\%} & \textbf{845M} \\ \hline
ResNet\_101 & 23.6\% & 7.1\%& 7600M \\
\bf{Our Model \textbf{O}} & \textbf{23.99\%}& \textbf{7.12\%}& \textbf{1172M} \\ \hline
\end{tabular}
\end{center}
\caption{ Top-1 and Top-5 error rate of single-crop testing with single model, number of multiplication of our model and several previous work. The numbers in this table are generated with single model and center-crop. For AlexNet and GoogLeNet, the top-1 error is missing in original paper and we use the number of Caffe's implementation\cite{caffe}. For ResNet-34, we use the number with Facebook's implementation\cite{facebook-torch}.}
\label{tab: comp-state-of-the-art}
\end{table*}

\begin{figure*}[t]
\centering
\subfigure[$3 \times 3$ standard convolutional layer] {
  \includegraphics[width=0.2\linewidth ]{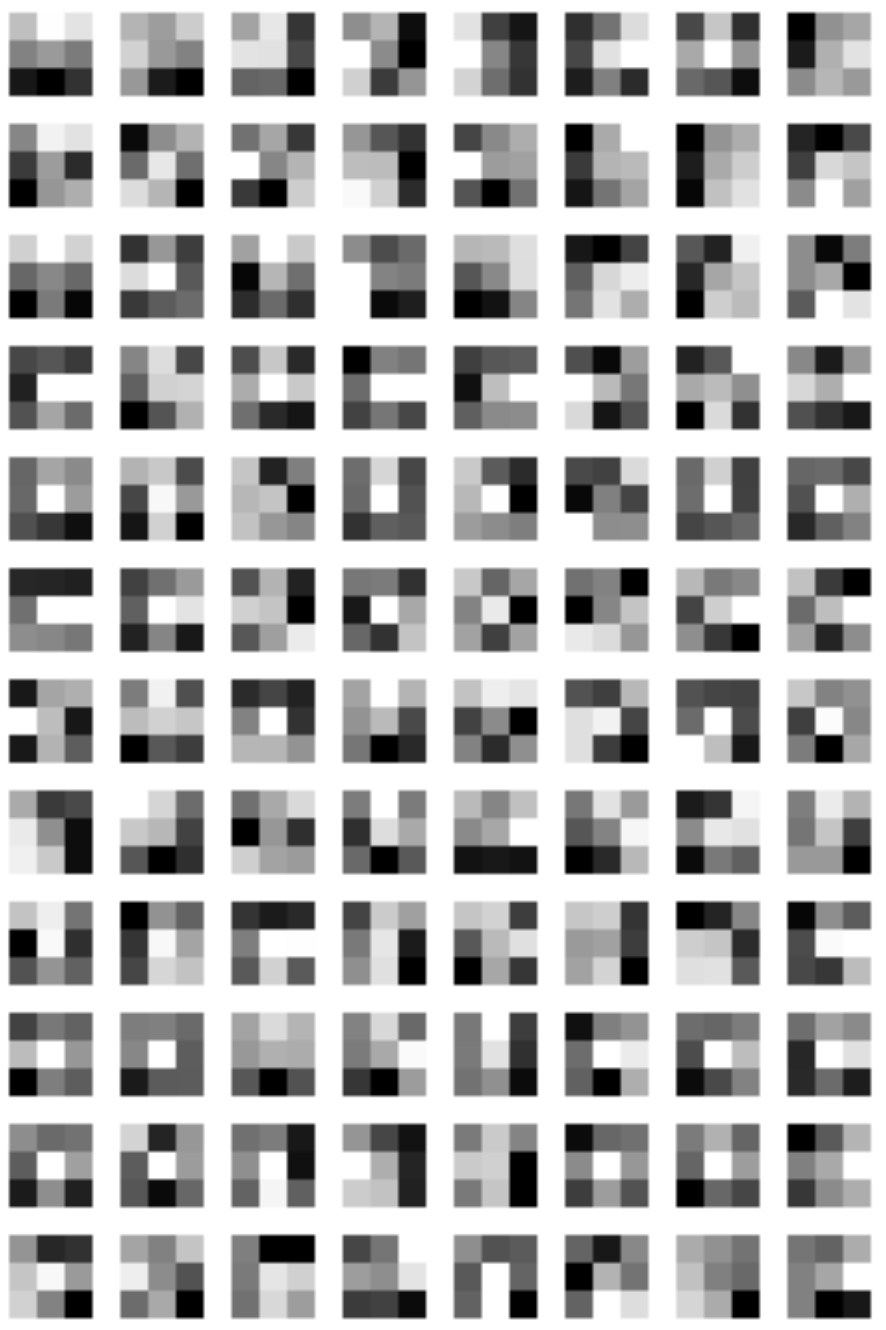}
  \label{fig: 3x3-baseline-filter}
}
\subfigure[$3 \times 3$ single intra-channel convolutional layer] {
  \includegraphics[width=0.2\linewidth]{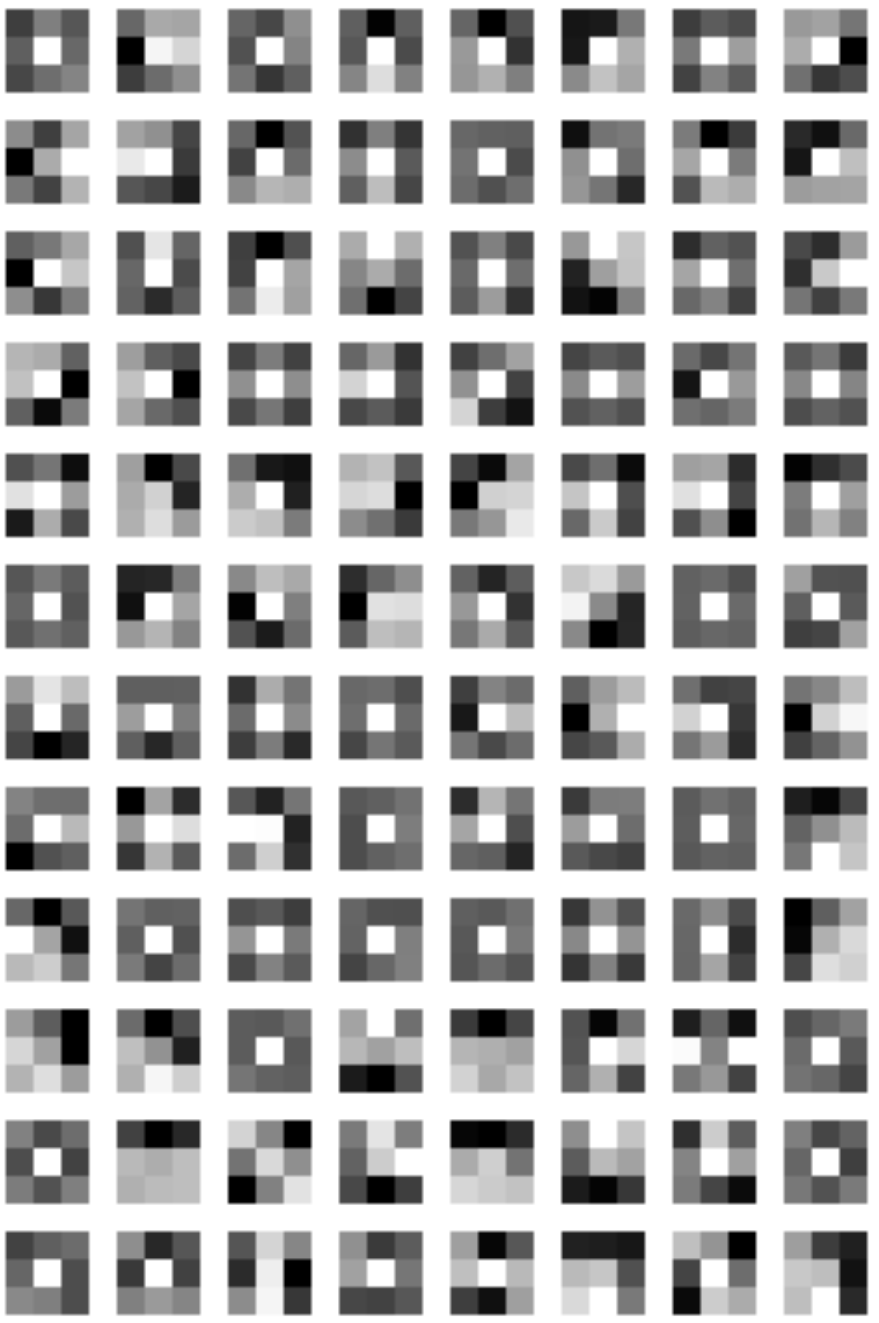}
  \label{fig: 3x3-filter}
}
\subfigure[$5 \times 5$ standard convolutional layer] {
  \includegraphics[width=0.2\linewidth ]{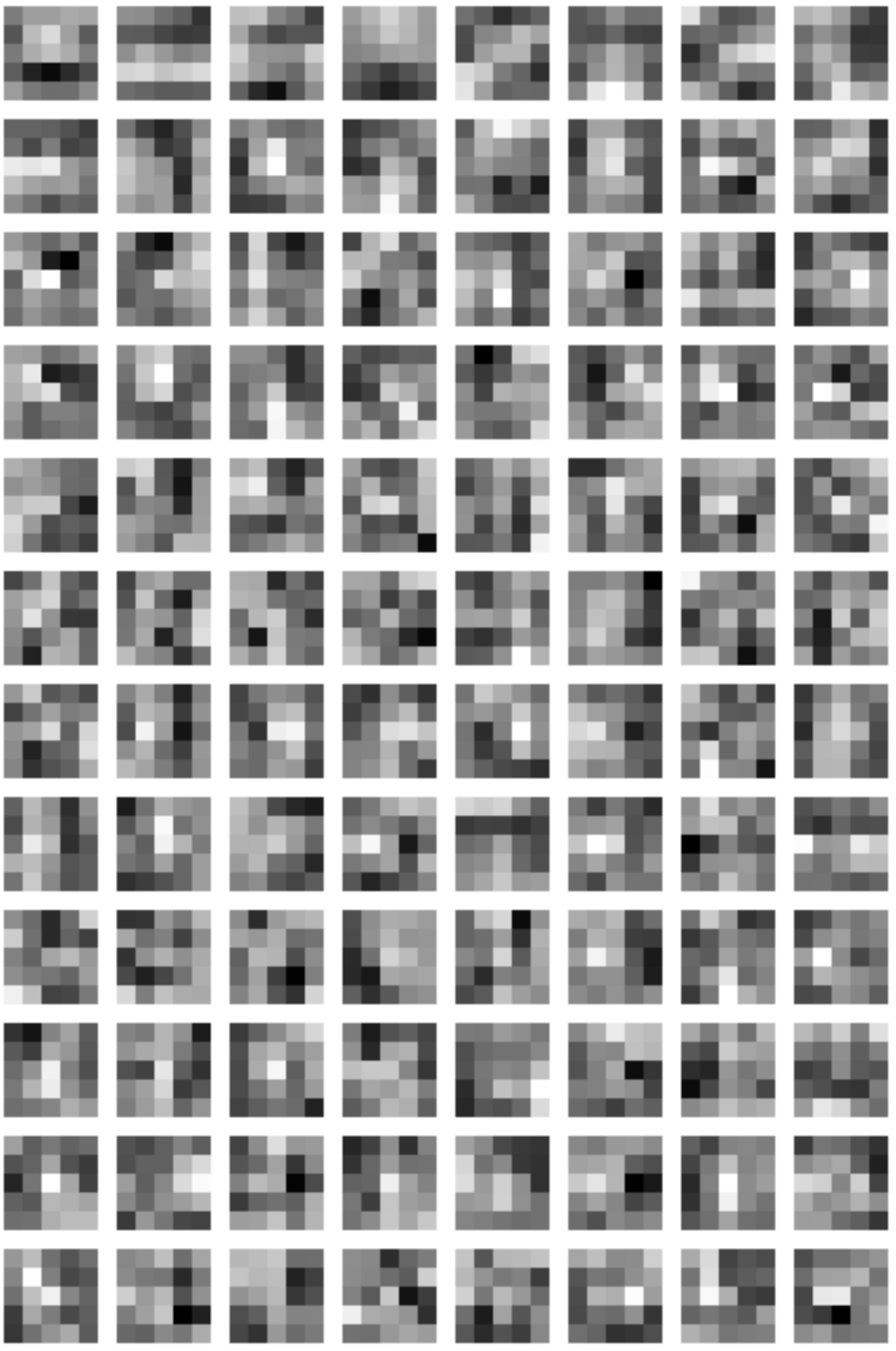}
  \label{fig: 5x5baseline-filter}
}
\subfigure[$5 \times 5$ single intra-channel convolutional layer] {
  \includegraphics[width=0.2\linewidth]{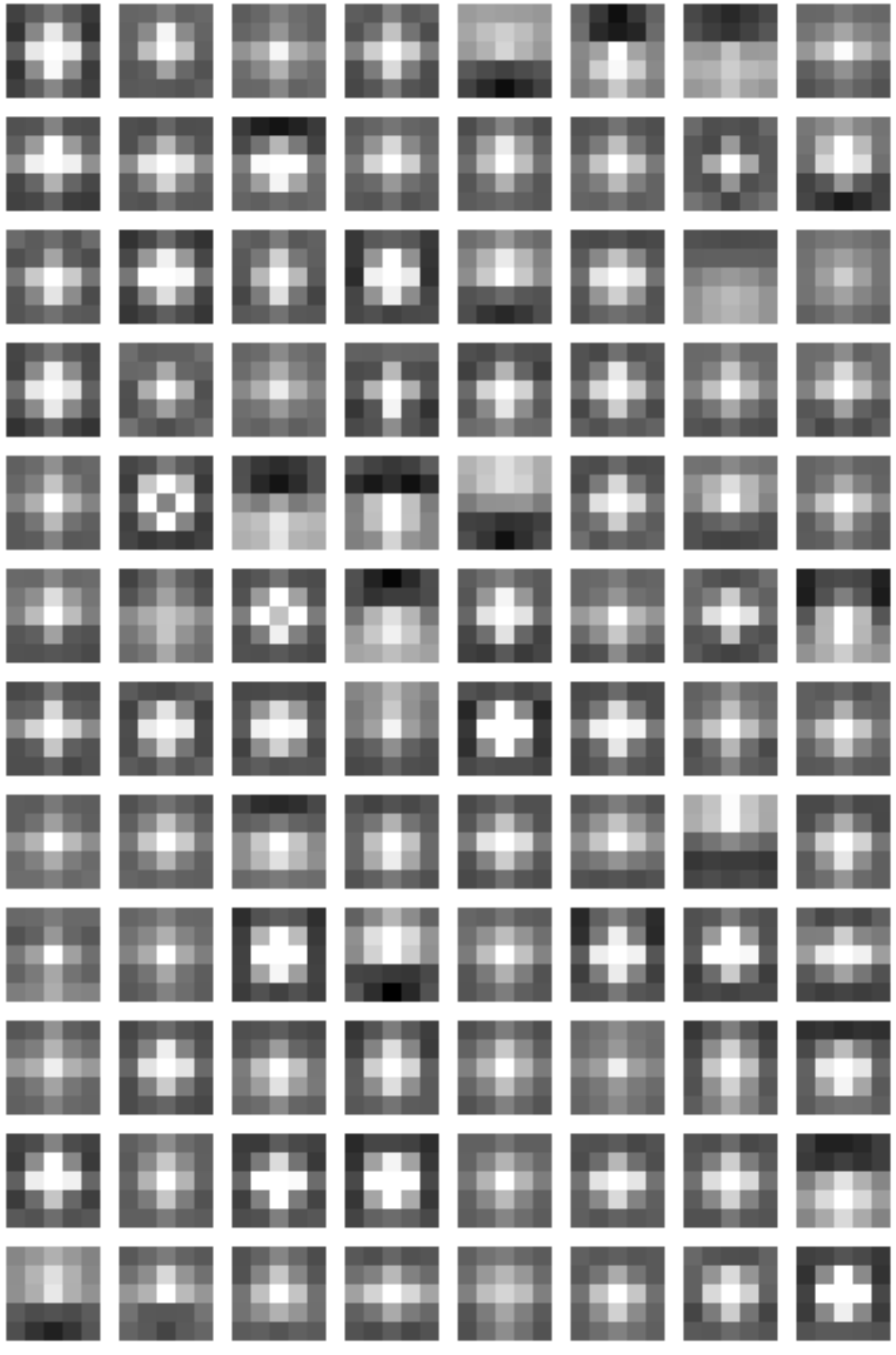}
  \label{fig: 5x5-filter}
}
 \caption{Visualization of convolutional kernels. We compare the $3\times 3$ and $5\times 5$ kernels that are learned by the proposed single intra-channel convolutional layer and the standard convolutional layer. The kernels from single intra-channel convolution exhibit a higher level of regularity in structure.}
 \label{fig: visualization}
\end{figure*}

Figure \ref{fig: scatter} compares the accuracy and complexity of our model from $\bf{L}$ to $\bf{O}$ with several previous works. Table \ref{tab: comp-state-of-the-art} lists the detailed results.  
 Figure \ref{fig: scatter} provides a visual comparison in the form of scattered plot. The red marks in the figure represent our models. All of our models demonstrate remarkably lower complexity while being as accurate. Compared to VGG, Resnet-34, Resnet-50 and Resnet-101 models, our models are $\bf{42\times}$, $\bf{7.3\times}$, $\bf{4.5\times}$, $\bf{6.5\times}$ more efficient respectively with similar or lower top-1 or top-5 error.

\subsection{Visualization of filters}

Given the exceptionally good performance of the proposed methods, one might wonder what type of kernels are actually learned and how they compare with the ones in traditional convolutional layers. We randomly chose some kernels in the single intra-channel convolutional layers and the traditional convolutional layers, and visualize them side by side in Figure \ref{fig: visualization} to make an intuitive comparison. Both $3\times 3$ kernels and $5\times 5$ kernels are shown in the figure. The kernels learned by the proposed method demonstrate much higher level of regularized structure, while the kernels in standard convolutional layers exhibit more randomness. We attribute this to the stronger regularization caused by the reduction of number of filters.

\subsection{Discussion on implementation details} 
In both SIC layer and spatial ``bottleneck'' structure , most of the computation is consumed by the linear channel projection, which is basically a matrix multiplication. The 2D spatial convolution in each channel has similar complexity to a max pooling layer. Memory access takes most running time due to low amount of computation. The efficiency of our CUDA based implementation is similar to the open source libraries like Caffe and Torch. We believe higher efficiency can be easily achieved with an expert-level GPU implementation like in CuDNN. The topological subdivisioning layer resembles the structure of 2D and 3D convolution.Unlike the sparsity based methods, the regular connection pattern from topological subdivisioning makes efficient implementation possible. Currently, our implementation simply discards all the non-connected weights in a convolutional layer.

\section{Conclusion}
This work introduces a novel design of efficient convolutional layer in deep CNN that involves three specific improvements: (i) a single intra-channel convolutional (SIC) layer ; (ii) a topological subdivision scheme; and (iii) a spatial ``bottleneck'' structure. As we demonstrated, they are all powerful schemes in different ways to yield a new design of the convolutional layer that has higher efficiency, while achieving equal or better accuracy compared to classical designs. While the numbers of input and output channels remain the same as in the classical models, both the convolutions and the number of connections can be optimized against accuracy in our model - (i) reduces complexity by unraveling convolution, (ii) uses topology to make connections in the convolutional layer sparse, while maintaining local regularity and (iii) uses a conv-deconv bottleneck to reduce convolution while maintaining resolution. Although the CNN have been exceptionally successful regarding the recognition accuracy, it is still not clear what architecture is optimal and learns the visual information most effectively. The methods presented herein attempt to answer this question by focusing on improving the efficiency of the convolutional layer. We believe this work will inspire more comprehensive studies in the direction of optimizing convolutional layers in deep CNN. 

\bibliographystyle{unsrt}
\bibliography{egbib}

\end{document}